\pdfoutput=1

\documentclass[11pt]{article}

\usepackage{amsmath}
\usepackage{amssymb}

\usepackage[final]{acl}

\usepackage{times}
\usepackage{latexsym}

\usepackage[T1]{fontenc}

\usepackage[utf8]{inputenc}

\usepackage{microtype}

\usepackage{inconsolata}

\usepackage{graphicx}

\usepackage{config}

\usepackage{booktabs}
\usepackage{arydshln}
\usepackage{xspace}
\usepackage{booktabs}

\usepackage{colortbl}
\usepackage{xcolor} 

\definecolor{lightgray}{gray}{0.9}

\title{
Can Compact Language Models Search Like Agents? Distillation-Guided\\
Policy Optimization for Preserving Agentic RAG Capabilities}

\author{
 \textbf{Rikuto Kotoge\textsuperscript{1}\thanks{Work done as a research intern at OMRON SINIC X.} }
 \textbf{Mai Nishimura\textsuperscript{2} }
 \textbf{Jiaxin Ma\textsuperscript{2}}
\\
\\
 \textsuperscript{1}The University of Osaka  
 \textsuperscript{2}OMRON SINIC X Corporation 
\\
 \small{
 \textsuperscript{1}\texttt{rikuto88@sanken.osaka-u.ac.jp  }
 \textsuperscript{2}\texttt{\{mai.nishimura, jiaxin.ma\}@sinicx.com }
 }
}

\begin{document}
\maketitle

\begin{abstract}
Reinforcement Learning has emerged as a \diff{dominant} post-training approach to elicit agentic RAG behaviors such as search and planning from language models.
\diff{Despite its success with larger models, applying RL to compact models (\eg 0.5--1B parameters) presents unique challenges.
The compact models exhibit poor initial performance, resulting in sparse rewards and unstable training.}
To overcome these difficulties, we propose Distillation-Guided Policy Optimization (DGPO), which 
\diff{employs}
cold-start initialization from teacher demonstrations and continuous teacher guidance during policy optimization. 
\diff{To understand how compact models preserve agentic behavior, }
we introduce Agentic RAG Capabilities (ARCap), a fine-grained metric analyzing reasoning, search coordination, and response synthesis.
Comprehensive experiments demonstrate that DGPO enables compact models to achieve sophisticated agentic search behaviors, even outperforming the larger teacher model in some cases. DGPO makes agentic RAG feasible in computing resource-constrained environments.

\vspace{0.5em}
\noindent
{\small
\faGlobe~\textbf{Project}~\href{https://omron-sinicx.github.io/dgpo}{\texttt{omron-sinicx.github.io/dgpo}}\\
\faGithub~\textbf{Code}~\href{https://github.com/omron-sinicx/dgpo}{\texttt{omron-sinicx/dgpo}}\\
\faDatabase~\textbf{Models}~\href{https://huggingface.co/collections/omron-sinicx/dgpo}{\texttt{omron-sinicx/dgpo}}
}
\end{abstract}

\section{Introduction}
\label{sec:intro}
\begin{figure}[t]
  \centering
  \includegraphics[width=\linewidth]{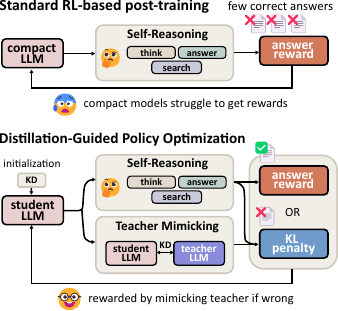}
  \caption{\textbf{Distillation-Guided Policy Optimization.}
  Top: Compact models struggle to earn rewards due to poor capability, which leads to training collapse.
  Bottom: \method establishes a stable reward mechanism by guiding incorrect answers through teacher mimicry.
  }
  \label{fig:distillation-guided-rl}
\end{figure}
\diff{Agentic RAG~\cite{agenticrag} has emerged as a new paradigm where LLMs function as autonomous search agents, coordinating retrieval, query reformulation, and evidence integration.}
\diff{While externalizing knowledge storage, these systems require}
sophisticated reasoning abilities within the LLMs for effective search coordination.
Consequently, existing agentic RAG systems predominantly rely on large language models with billions of parameters~\cite{xu2025comprehensivesurveydeepresearch}, 
leaving the potential of agentic RAG in resource-constrained environments largely unexplored.
The emergence of small language models (SLMs)~\citep{slmagent}, particularly compact models (e.g., 0.5--1B)
raises a compelling question:
\emph{can we unlock the latent potential of compact language models to acquire the art of agentic RAG?}

\diff{Eliciting agentic search capabilities from smaller language models typically requires two approaches:}
reinforcement learning (RL) via self-exploration and knowledge distillation (KD) from a teacher model. We refer to the compact model under training as the \emph{student}, regardless of the approach. Yet both approaches
\diff{become largely ineffective for compact models (0.5--1B) due to their poor initial capabilty.}
\diff{RL~\cite{ppo,grpo} suffers from sparse rewards and poor exploration due to weak student-generated outputs (SGOs).}
\diff{Standard KD}~\cite{kd,shing2025taid} 
using only teacher-generated outputs (TGOs) leads to exposure bias \cite{NIPS2015_e995f98d} while on-policy distillation methods \cite{gu2024minillm,gkd} also suffer from the noisy and low-quality nature of SGOs. 
\diff{Neither approach addresses the fundamental bottleneck of poor initial output quality in compact models.}

\diff{To overcome this fundamental bottleneck,}
we propose Distillation-Guided Policy Optimization (DGPO), a novel RL framework that addresses the core issue of low-quality SGOs through the strategic integration of teacher guidance and RL (\Cref{fig:distillation-guided-rl}).
DGPO operates through two key mechanisms. First, cold-start initialization \diff{through KD using TGOs} dramatically stabilizes early training by providing high-quality initial trajectories. 
\diff{Second, selective teacher guidance during RL that rewards correct self-reasoning while providing teacher mimicry for incorrect attempts.}
\diffJ{This synergy between selective KL-based teacher guidance and RL-driven self-exploration allows the compact model to discover policies that outperform the teacher 
in some experimental settings.}

\begin{figure}[t]
  \centering
  \includegraphics[width=\linewidth]{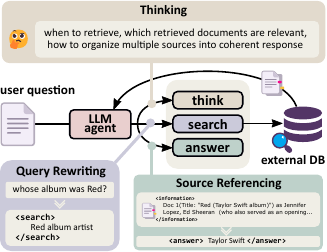}
  \caption{\textbf{Agentic RAG capability.} We introduce Agentic RAG Capability (ARCap) which characterizes the core capabilities of LLMs required for agentic RAG systems. ARCap is evaluated as three primary components: \emph{thinking}, \emph{query rewriting}, and \emph{source referencing}. }
  \label{fig:arc}
\end{figure}

\diff{To understand how DGPO preserves agentic capability in compact models,}
we introduce Agentic RAG Capabilities (ARCap), a fine-grained evaluation framework that decomposes the agentic search into three core dimensions: \diff{\emph{thinking}, \emph{query rewriting}, and \emph{source referencing} (Figure~\ref{fig:arc}).
Unlike conventional metrics that focus on final accuracy, ARCap evaluates the agentic search process, revealing}
how different aspects of agentic behavior emerge and decline across different models. 
\diff{Comprehensive evaluations} demonstrate that DGPO consistently outperforms baselines in \diff{final accuracy}.
ARCap reveals that DGPO improves multi-hop reasoning and coordination while maintaining competitive performance in source referencing and query rewriting. 
\diff{Such capability-level insights are crucial for advancing agentic RAG in compact models.}

\diff{Our contributions are summarized in four key dimensions.
\textbf{(i) Problem:} we pioneer the challenging domain of agentic RAG post-training for extremely compact models (0.5--1B), identifying fundamental challenges that existing methods fail to address.
\textbf{(ii) Methodology:} We propose Distillation-Guided Policy Optimization (DGPO), an RL framework designed to stabilize training in compact models via cold-start initialization and selective teacher guidance. 
\textbf{(iii) Evaluation:} 
we present ARCap, a capability-level evaluation framework that provides a detailed diagnosis of agentic behavior.
\textbf{(iv) Results:}
DGPO outperforms RL and distillation baselines
\diffJ{across multiple model families and sizes.}
Remarkably, our method achieves \textbf{teacher-surpassing performance} on several datasets. 
}

\section{Related Work}
\label{sec:related}
\paragraph{Agentic RAG.}

WebGPT~\cite{webgpt} introduced RLHF-driven browser interaction for retrieval-grounded QA.  
ReAct~\cite{yao2023react} generalized this idea by interleaving chain-of-thought and tool calls via special \texttt{<think>} or \texttt{<act>} tokens.
To tighten the coupling between retrieval and reasoning, IRCoT \cite{ircot} explicitly alternates each chain-of-thought (CoT) step with a retrieval.
Adaptive-RAG~\cite{adaptiverag} further predicts retrieval steps based on question complexity.
Most recently, Search-R1 \cite{searchr1} leveraged RL to teach an LLM to generate multi-turn search queries, achieving state-of-the-art results.
Our work specifically focuses on enabling agentic RAG in compact models and introduces a comprehensive evaluation framework for multi-dimensional capability evaluation.

\paragraph{Post-training for LLMs.}
RL algorithms such as PPO \cite{ppo} and GRPO \cite{grpo} have proven effective in enhancing reasoning capabilities for LLMs \cite{comanici2025gemini25pushingfrontier, yang2025qwen3technicalreport}, particularly in domains like mathematical problem solving. 
At the initial stage of training, base models require sufficient performance to obtain meaningful rewards; otherwise, sparse reward signals lead to training instability. To address this cold-start problem, DeepSeek-R1~\cite{deepseekai2025deepseekr1incentivizingreasoningcapability} demonstrates that SFT-based model initialization effectively warms up the model prior to RL, achieving favorable results through CoT demonstrations.
Our work is the first to integrate distillation principles into both cold-start initialization and concurrent RL training, enabling stable distillation-guided learning in compact models.

\paragraph{Knowledge Distillation for LLMs.}
Knowledge distillation (KD) \citep{kd} enables smaller student models to learn from larger teacher models by matching softened output distributions.
To mitigate the capacity gap between student and teacher models \cite{Mirzadeh_Farajtabar_Li_Levine_Matsukawa_Ghasemzadeh_2020, zhang-etal-2023-lifting}, some methods use interpolated or smoothed intermediate student and teacher distributions \cite{distillm, shing2025taid}.
However, because these methods rely on TGOs during training while inference still uses the SGOs, a train–inference mismatch arises, leading to exposure bias \cite{NIPS2015_e995f98d}. 
To mitigate this, recent work also proposes on-policy distillation from SGOs \cite{gkd, gu2024minillm, yang2025qwen3technicalreport}, where the student learns directly from its own generated outputs during training.
Another limitation is that distillation methods require sensitive teacher-guidance schedulers \cite{distillm,shing2025taid}, which are hard to tune and affect stability.
Meanwhile, conventional single-hop RAG distillation methods~\citep{kard,reaugkd,Liao_He_Xu_Zhang_Liu_Zhao_2025} typically use sequence-level knowledge distillation (SeqKD) \citep{seqkd}, an SFT-based approach where students learn from teacher-generated CoT demonstrations rather than matching output distributions.
 While SeqKD improves the reasoning capability of SLMs~\citep{deepseekai2025deepseekr1incentivizingreasoningcapability}, 
cross-entropy training with hard targets alone is insufficient to transfer the richer distributional knowledge from the teacher \citep{kd}.
 Our method addresses these challenges through a two-stage approach: off-policy KD on TGOs for cold-start initialization, followed by off-policy KD-driven RL optimization on SGOs to further extract model capabilities without requiring handcrafted schedulers.

\section{Agentic RAG}
\label{sec:pre}
\begin{table}[t]
   \begingroup
   \tabcolsep = 1.5mm
   \small
   \begin{tabularx}{\linewidth}{p{7.5cm}}
     \toprule[1.2pt]
     \textbf{Question:} Whose album was Red?\\
     \midrule[1.2pt]
     \colortag{olive}{think}To answer this question, I need to find out which artist released the album titled "Red".\colortag{olive}{/think}\\
     \midrule
    \colortag{sblue}{search} Red album artist \colortag{blue}{/search}\\
    \midrule
    \colortag{brown}{information}\textbf{Doc 1(Title: "Red (Taylor Swift album)")} as Jennifer Lopez, Ed Sheeran (who also served as an \dots \\ 
    \textbf{Doc 2(Title: "Red (band)")} deal with Christian label Essential Records in 2004. Their debut albums \dots \\ 
    \textbf{Doc 3(Title: "Red (Dia Frampton album)")} Red (Dia Frampton album) Red is the debut  \dots 
    \colortag{brown}{/information} \\
    \midrule
    \colortag{olive}{think}
    From the information provided, it is clear that the album "Red" was released by Taylor Swift. It was her fourth studio album and had a country sound. 
    \colortag{olive}{/think}
    \\
    \midrule
\colortag{teal}{answer} Taylor Swift \colortag{teal}{/answer}\\
\bottomrule[1.2pt]
   \end{tabularx}
   \endgroup
   \caption{\textbf{An example of Reasoning Processes.} The agent exhibits \emph{thinking} through deliberative reasoning, \emph{query rewriting} by reformulating the original query into the actual search query, and \emph{source referencing} by incorporating retrieved information in to the final answer.} 
\label{tab:reasoning-example}
\end{table}

\subsection{Framework}
While agentic RAG~\cite{agenticrag} encompasses diverse interpretations, we define it through two key characteristics: (1) autonomous, adaptive search coordination behaviors, and (2) three fundamental actions—\emph{thinking}, \emph{searching}, and \emph{answering}—that constitute agentic intelligence.
In agentic RAG, LLMs function as agents making sequential decisions at each timestep $t$. 
Given a user question $\bm x$ and an external retrieval system $\mathcal R$, 
the LLM agent operates as a policy 
$\pi_\theta(\bm y_t | \bm x_t; \mathcal R)$
, where
\[
  \bm y \!\in
  \!\Bigl\{\;
      \underbrace{\textsc{Think}(\cdot)}_{\text{reasoning token}},\;
      \underbrace{\textsc{Search}(\cdot)}_{\text{search query}},\;      
      \underbrace{\textsc{Answer}(\cdot)}_{\text{forming an answer}}
  \Bigr\}\,.
\]
As demonstrated in \Cref{tab:reasoning-example}, we employ structured tokens~\cite{searchr1} to organize the actions: \colortag{olive}{think} for reasoning, \colortag{sblue}{search} for database queries, \colortag{brown}{information} for retrieved documents, and \colortag{teal}{answer} for final responses.
\final{The model can invoke \colortag{olive}{think} and \colortag{sblue}{search} at arbitrary times and for an arbitrary number of steps, while \colortag{teal}{answer} is used exactly once at the end to produce the final output.}

\subsection{Agentic RAG Capability (ARCap)}
We propose Agentic RAG Capability (ARCap) as a comprehensive metric to systematically evaluate agentic behavior across multiple dimensions. As demonstrated in \Cref{tab:reasoning-example}, we characterize ARCap through three core dimensions:

\paragraph{Source Referencing.}
Accurately incorporating retrieved information into final answers (shown in the \colortag{brown}{information} and \colortag{teal}{answer} entries).

\paragraph{Query Rewriting.}
Reformulating user questions into effective search queries, as literal keyword matching often fails to retrieve relevant documents. The agent must paraphrase key concepts and introduce related terms to maximize retrieval effectiveness (illustrated by transforming "Whose album was Red?" into "Red album artist" in \colortag{sblue}{search}).

\paragraph{Thinking.}
Making informed decisions about when to retrieve information, which documents contain relevant answers, and how to synthesize multiple pieces of evidence into coherent responses. This involves assessing context sufficiency and integrating retrieved sources in a logically consistent manner (demonstrated in \colortag{olive}{think} entries).

\begin{figure}[t]
  \centering
  \includegraphics[width=\linewidth]{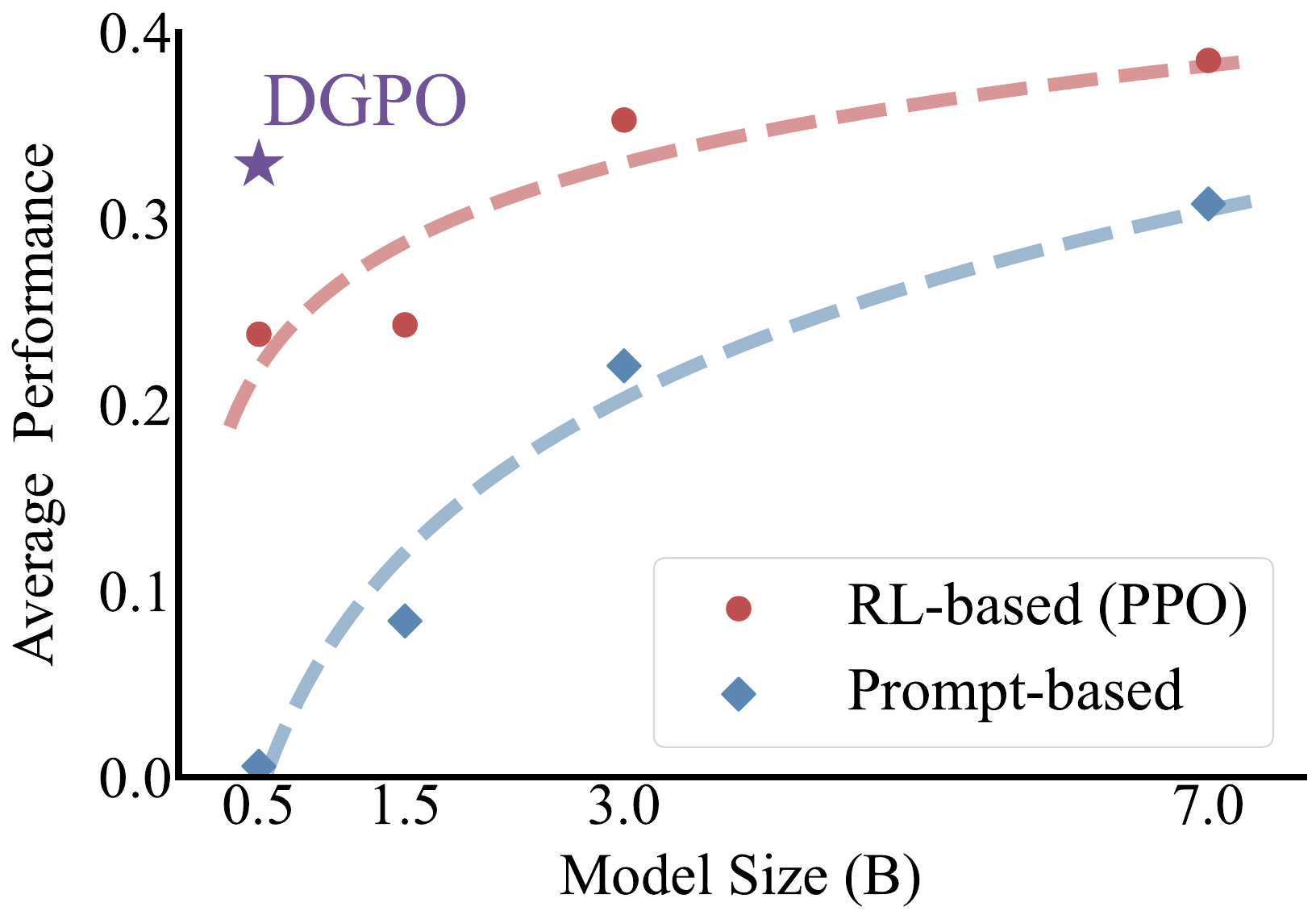}
  \caption{Comparison of prompt-based and RL-based (PPO) post-training agentic RAG across model sizes.}
  \label{fig:scaling}
\end{figure}

\begin{figure}[t]
  \centering
  \includegraphics[width=\linewidth]{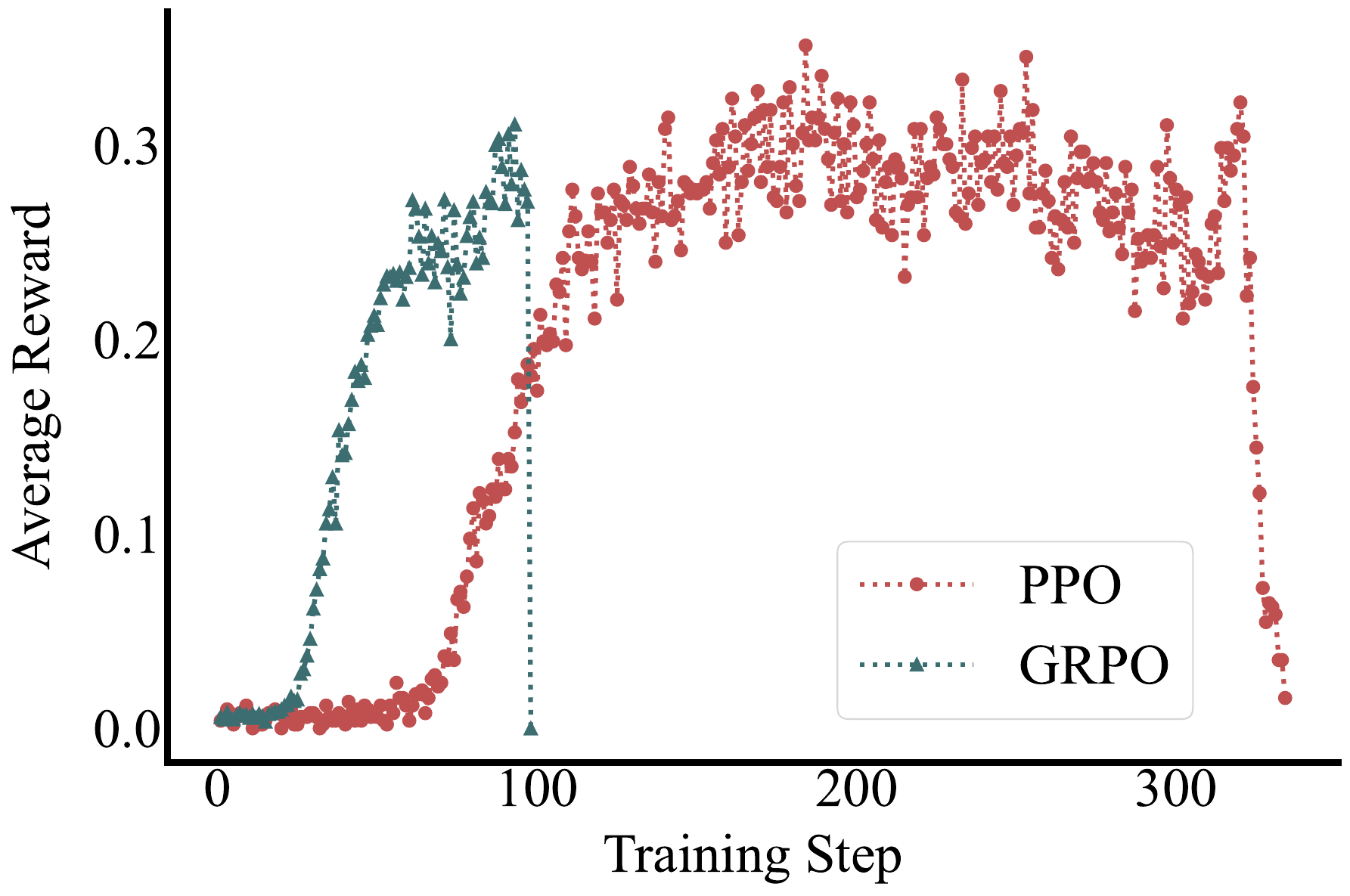}
  \caption{\final{Training curve of PPO and GRPO.}}
  \label{fig:limitation}
\end{figure}

\begin{figure*}[t]
  \centering
  \includegraphics[width=0.9\linewidth]{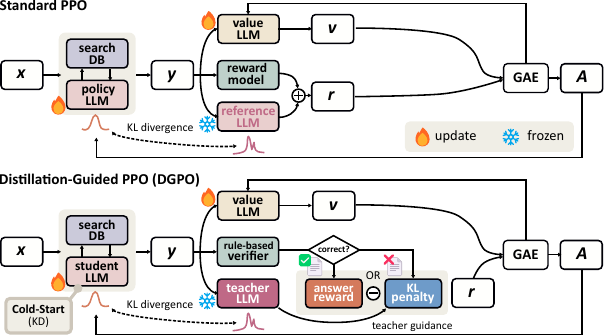}
  \caption{
  Top: Standard PPO pipeline for post-training LLMs. The reference LLM serves as a regularization anchor to prevent excessive deviation from the initial policy.
Bottom: Our proposed distillation-guided PPO pipeline. Unlike conventional approaches where the reference model merely constrains policy drift, our framework employs the teacher model to actively guide the student toward correct behaviors when autonomous attempts fail, transforming the reference's role from passive regularization to active pedagogical guidance.}
  \label{fig:dgpo}
\end{figure*}

\subsection{Challenges in Compact Models.}
\paragraph{\final{Performance Gap.}}
Our preliminary experiments compared the performance of prompt-based and RL-based agentic RAG models across various model sizes, evaluated on the average of seven QA datasets (Figure~\ref{fig:scaling}). Here prompt-based refers to Qwen2.5-instruction checkpoints and RL-based refers to post-trained models using PPO~\cite{searchr1} tailored for agentic RAG. The experimental setup is detailed in Section~\ref{sec:exp}. While RL models boosted performance overall in the context of agentic RAG, smaller models still lagged far behind their larger counterparts. 
We include this result here to highlight the limitations of applying RL directly to compact models—an observation that motivates our proposed approach, \method, introduced in the next section.

\paragraph{\final{Training Instability.}}
\final{
Figure~\ref{fig:limitation} presents the RL training curves of Qwen2.5-0.5B-instrtuct model with PPO \citep{ppo} and GRPO \citep{grpo} for agentic RAG. 
Smaller models converge faster but tends to become unstable relatively early in training \cite{searchr1}, preventing further performance gains beyond that point.
PPO provides more stable optimization than GRPO but converges slower.
}

\section{DGPO: Distillation-Guided Policy Optimization}
\label{sec:pre}

\subsection{Core Framework}
\Cref{fig:dgpo} depicts our framework which  combines distillation and reinforcement learning to train compact agentic RAG models through a two-phase learning strategy, eliminating the need for a handcrafted scheduler. Early-stage student-generated outputs (SGOs) are often noisy and unstable, while teacher-generated outputs (TGOs) provide quality guidance but suffer from exposure bias. To address these challenges, we propose two key mechanisms:
\paragraph{Cold-Start Initialization via KD.}
In the initial phase, students learn purely from TGOs via knowledge distillation. This provides stable, high-quality trajectories that dramatically improve early training dynamics and establish a strong foundation for subsequent RL optimization.
\paragraph{Selective KL penalty.}
During the RL phase, we apply KL divergence penalties selectively—only to incorrect predictions—guiding students toward informative teacher behaviors while preserving exploration capabilities. This targeted regularization enables autonomous reasoning development without being overly constrained by the teacher model.

\subsection{KD initialization with TGOs}
During the cold-start phase, we initialize the student model by distilling from a strong teacher policy using a general KD loss that combines cross-entropy from hard labels and KL divergence as: 
\begin{equation}
\resizebox{\linewidth}{!}{%
$
\begin{aligned}
\mathcal{L}_{\text{distill}}\!\! =\!\! \mathcal{L}_{\text{CE}}(\pi_{\text{g}}, \pi_\theta)\!\! +\!\! \lambda D_{\text{KL}}\!\!\left[\pi_{\text{g}}(\cdot\!\! \mid\!\! x) \| \pi_\theta(\cdot \!\!\mid \!\!x)\right]\,,
\end{aligned}
$
}
\end{equation}
where $\pi_\theta$ denotes the student policy and $\pi_{\text{g}}$ is the frozen teacher. 
We filter TGOs to retain only correct outputs, ensuring the student $\pi_{\theta}$
learns from high-quality teacher samples. 

\subsection{Distillation-guided RL with SGOs}
Upon reaching a performance threshold, we transition to PPO-based RL using the distilled student as the initial policy. This staged approach stabilizes training dynamics and improves sample efficiency, particularly when the student model has significantly fewer parameters than the teacher. By avoiding premature exploration from a weak policy, our method ensures that RL begins with a reasonable approximation of agentic behaviors.

\paragraph{PPO with Search Engine}

Proximal Policy Optimization (PPO)~\cite{ppo} is a widely used RL algorithm for LLM fine-tuning, offering stable training for compact models.
Our method optimizes LLMs with search engine $\se$ by maximizing the following objective,
\begin{equation}
\resizebox{\linewidth}{!}{
\ensuremath{                
\begin{aligned}
  \mathbb{E}_{x \sim \mathcal{D},
             y \sim \pi_{\text{old}}(\!\cdot\mid x;\se)}\!
  &\Biggl[
      \frac{1}{\sum_{t=1}^{|y|}\! \mathbbm{1}(y_t)}
      \!\!\!\!\!
      \sum_{\substack{t=1\\ \mathbbm{1}(y_t)=1}}^{|y|}
      \!\!\!\!
      \min\!\Bigl(
        \frac{\pi_{\theta}(y_t \mid x, y_{<t}; \se)}
             {\pi_{\text{old}}(y_t \mid x, y_{<t}; \se)}
        A_t,\, \\
        &\operatorname{clip}\Bigl(
          \frac{\pi_{\theta}(y_t\!\! \mid\!\!x, y_{<t}; \se)}
               {\pi_{\text{old}}(y_t\!\! \mid \!\!x, y_{<t}; \se)},\!
          1\!\!-\!\!\epsilon,\,1\!\!+\!\!\epsilon
        \Bigr)A_t
      \Bigr)
    \Biggr],
    \label{eq:ppo}
\end{aligned}
}
}
\end{equation}
where \( \pi_{\theta} \) and \( \pi_{\text{old}} \) represent the current and previous student policy models, respectively. 
\( x \) denotes input samples
and \( y \) represent the generated outputs interleaved with search engine calling results. 
The term \( \epsilon \) is a clipping-related hyperparameter introduced in PPO to stabilize training. 
The advantage estimate \( A_t \) is computed using Generalized Advantage Estimation (GAE) \citep{gae}, based on future rewards and a learned value function.
$\mathbbm{1}(y_t)$ is a token loss masking operation.
See \cref{sec:token-masking} for details on token masking.

\paragraph{Reward and Selective KL penalty}
We employ binary exact matching (EM) for answer rewards to prevent reward hacking:
\begin{equation}
r_{\text{answer}}(x, y) = 
\begin{cases*}
1 & if  $y = y^*$ \\
0 & otherwise\,,
\end{cases*}
\label{eq:reward}
\end{equation}
where $y$ is the predicted answer and $y^*$ is the ground-truth.
However, \cref{eq:reward} provides no learning signal for incorrect predictions, 
causing training stagnation with poor SGOs.
To address this, we introduce selective KL penalty. The student $\pi_{\theta}$ receives reward for correct self-reasoning, but when incorrect, the teacher $\pi_\text{g}$ guides the student to mimic teacher behavior through KL regularization,
\begin{equation}
\resizebox{\linewidth}{!}{%
$
\begin{aligned}
r_\phi(x, y) = 
\begin{cases}
1 & \text{if } y = y^* \\
- \beta D_{\text{KL}}\left[ \pi_\theta(y \!\mid \!\!x; \!\!\mathcal R) \|  \pi_{\text{g}}(y\!\mid \!\!x; \!\mathcal R) \right] & \text{otherwise}.
\end{cases}
\end{aligned}
$
}
\end{equation}
As illustrated in \Cref{fig:dgpo}, our approach differs fundamentally from standard PPO-based LLM tuning. While conventional PPO uses a frozen initial LLM as a reference regularizer to prevent excessive drift from the initial policy, DGPO employs the teacher LLM as an active guide that steers the student toward correct behaviors when errors occur. This can be seen as a form of targeted regularization~\cite{pmlr-v97-laroche19a}, which allows free exploration during correct predictions but applies corrective guidance through KL penalties when the student fails. By selectively emphasizing high-divergence incorrect outputs, our method focuses learning on error correction while maintaining autonomous reasoning capabilities, resulting in efficient and stable training.

\section{Experiments}
\label{sec:exp}

\begin{table*}[t]
\centering

\resizebox{\textwidth}{!}{%
\begin{tabular}{l|ccc|cccc|c}
\toprule
\textbf{\colorbox{blue}{Qwen 2.5 (3B $\rightarrow$ 0.5B)}} & NQ & TriviaQA & PopQA & HotpotQA & 2wiki & MuSiQue & Bamboogle & Avg. \\
\midrule
\textcolor{gray}{Student-0.5B}        & \textcolor{gray}{0.004} & \textcolor{gray}{0.006} & \textcolor{gray}{0.007} & \textcolor{gray}{0.007} & \textcolor{gray}{0.015} & \textcolor{gray}{0.000} & \textcolor{gray}{0.000} & \textcolor{gray}{0.006} \\
\textcolor{gray}{Teacher-3B}          & \textcolor{gray}{0.365} & \textcolor{gray}{0.569} & \textcolor{gray}{0.393} & \textcolor{gray}{0.340} & \textcolor{gray}{0.368} & \textcolor{gray}{0.135} & \textcolor{gray}{0.298} & \textcolor{gray}{0.353} \\
\hdashline
PPO \cite{searchr1} & 0.306 & \colorbox{yellow}{0.444} & \colorbox{yellow}{0.379} & 0.205 & 0.218 & 0.041 & 0.073 & 0.238 \\
\final{SFT $\rightarrow$ PPO} & 0.338 & 0.415 & 0.359 & 0.296 & 0.275 & 0.088 & 0.250 & 0.289 \\
GKD \cite{gkd}               &  0.266&   0.408&  0.358& 0.216& 0.217&  0.055&  0.161& 0.240 \\
SeqKD \cite{seqkd}                  & 0.331 & 0.416 & 0.364 & 0.283 & 0.273 & 0.089 & 0.169 & 0.275 \\
KD \cite{kd}               &  0.331& 0.431&  0.373&  0.286& \colorbox{yellow}{0.284} & 0.091 & \colorbox{green!80}{0.290} & \colorbox{yellow}{0.298} \\
DistiLLM \cite{distillm}               & \colorbox{yellow}{0.333} & 0.442&  0.373&  0.288&  0.270&  \colorbox{yellow}{0.095}&  0.209& 0.287 \\
TAID \cite{shing2025taid}                & 0.325 & 0.427&  0.365&  \colorbox{yellow}{0.290}& 0.270 & 0.079 & 0.218 & 0.282 \\
\midrule
\textbf{\method} (ours)               & \fcolorbox{black!60}{green!80}{\textbf{0.378}} &\colorbox{green!80}{0.481} &\fcolorbox{black!60}{green!80}{\textbf{0.402}}  &\fcolorbox{black!60}{green!80}{\textbf{{0.342}}}  &\colorbox{green!80}{0.303}  &\colorbox{green!80}{0.120}  &\colorbox{yellow}{0.274}  &\colorbox{green!80}{0.329}  \\
\bottomrule
\end{tabular}
}  
\caption{\diffJ{\textbf{Qwen 2.5 (3B $\rightarrow$ 0.5B) results} across different methods and QA benchmarks. The best and second-best results are highlighted in \colorbox{green!80}{green} and \colorbox{yellow}{yellow}, respectively. Teacher-surpassing scores are \fcolorbox{black!60}{green}{\textbf{bold and boxed}}.}}
\label{tab:qa_performance}
\end{table*}

\begin{table}[t]
\centering
\resizebox{0.8\linewidth}{!}{%
\begin{tabular}{lccc}
\toprule
Model family & \multicolumn{2}{c}{\colorbox{blue}{\textbf{Qwen 2.5}}} & \colorbox{red}{\textbf{Llama 3}} \\
Student size  & \multicolumn{2}{c}{0.5B}  & 1B \\
Teacher size  & 3B             & 7B  & 8B \\
\midrule
\textcolor{gray}{Student}    &\textcolor{gray}{0.006}     &  \textcolor{gray}{0.006}      & \textcolor{gray}{0.039}         \\
\textcolor{gray}{Teacher}  &  \textcolor{gray}{0.353}  & \textcolor{gray}{0.385}       &  \textcolor{gray}{0.438}        \\
\hdashline
PPO                 & 0.238 & 0.238       &   0.250      \\
KD                  & \colorbox{yellow}{0.298} & \colorbox{yellow}{0.280}       &  \colorbox{yellow}{0.347}       \\
\midrule
\textbf{\method}     & \colorbox{green}{\textbf{0.329}} & \colorbox{green}{\textbf{0.323}}       &     \colorbox{green}{\textbf{0.389}}    \\

\bottomrule
\end{tabular}
}
\caption{\diffJ{Average EM scores across seven QA benchmarks under different model configurations.}}
\label{tab:model_family}
\end{table}

\subsection{Experimental setup}
We focus our experiments on addressing the following questions:

\begin{description}[labelwidth=1em, leftmargin=2em, labelsep=.5em, align=left]
  \item[$\mathcal{Q}$1] Do our compact models preserve the overall performance of the teacher model?
  \item[$\mathcal{Q}$2] How well do compact models retain individual ARCap components? (a) \emph{Source Referencing}, (b) \emph{Query Rewriting}, (c) \emph{Thinking}.
  \item[$\mathcal{Q}$3] Which components of our method contribute most to performance improvements?
\final{
\item[$\mathcal{Q}$4] 
Does our method mitigate training instability in compact models?
}
\end{description}

\paragraph{Datasets.}
We evaluate \method on seven benchmark datasets, categorized as follows:
(1) General Question Answering: NQ \citep{nq}, TriviaQA \citep{triviaqa}, and PopQA \citep{popqa} datasets, which generally require single-hop searching, i.e., the answer can be derived from a single fact or passage.
(2) Multi-Hop Question Answering: HotpotQA \citep{yang2018hotpotqa}, 2WikiMultiHopQA \citep{2wiki}, MuSiQue \citep{musique}, and Bamboogle \citep{bamboogle} datasets, which require multi-hop searching over multiple evidence across different documents. \diff{Please See \cref{app:dataset-details} in details.}
\paragraph{Base Models.}
As the base student model, we use Qwen2.5-0.5B-instruct \cite{qwen25}. 
For the teacher model, we adopt Search-R1-PPO-3B based on Qwen2.5-3B-instruct.
\diff{
To assess generalizability across different model sizes and families, we also evaluate variants using Qwen2.5-7B-instruct and Llama 3 (Llama-3.2-1B-Instruct and Llama-3.1-8B-Instruct-based model) \citep{grattafiori2024llama3herdmodels}.
}
\paragraph{Baselines.}
We compare our method against baselines from three categories:
\begin{itemize}[nosep,leftmargin=*,topsep=0pt,partopsep=0pt,parsep=0pt,itemsep=0pt]
\item \emph{Reinforcement Learning:} Standard PPO~\cite{searchr1} illustrated in Figure~\ref{fig:dgpo} top
\footnote{We excluded GRPO~\cite{grpo} as it proved unstable for compact models, collapsing early due to poor SGOs.}.
\final{We consider two settings: PPO trained from scratch and PPO with a standard SFT warm start.}
\item \emph{On‑policy Distillation on SGOs:} GKD~\cite{gkd} 
minimizes reverse KL divergence between teacher and student distributions on SGOs.
\item \emph{Off‑policy Distillation on TGOs:} SeqKD \cite{seqkd} applies SFT on teacher outputs; KD~\cite{kd} combines cross-entropy loss with KL divergence; \diff{DistiLLM~\cite{distillm} 
adopts an adaptive off-policy strategy that integrates both SGOs and TGOs. 
TAID~\cite{shing2025taid} employs dynamic scheduling to interpolate from student to teacher distributions. Off-policy methods, except for DistiLLM,  train exclusively on correct TGOs\footnote{We observed that training on only the correct TGOs led to better performance.} }.
\end{itemize}
\diff{Detailed configurations for baseline and ablation variants can be found in Appendix~\ref{sec:ablation_baseline_setting}.
}

\paragraph{Evaluation Metrics.}
For all evaluations except the search results shown in Table \ref{tab:hit}, we use Exact Match (EM) as the evaluation metric, following \citet{searchr1,yu2024rankrag}.

\paragraph{Retrieval Settings.}
We follow \citet{searchr1} and  use the 2018 Wikipedia \cite{karpukhin-etal-2020-dense} as the knowledge source and E5 \cite{wang2024textembeddingsweaklysupervisedcontrastive} as
the retriever. 
We set the number of retrieved passages to 3.

\paragraph{Training Settings.}
We used the training sets of NQ and HotpotQA datasets.
Training was conducted on NVIDIA 8 × H200 GPUs.
Implementation details can be found in Appendix~\ref{app:implementation}.

\begin{table*}[t]
\centering
\begin{minipage}[t]{0.50\linewidth}
\centering
\resizebox{\linewidth}{!}{%
\begin{tabular}{lcccccc}
\toprule
Models & \multicolumn{2}{c}{NQ} & \multicolumn{2}{c}{MuSiQue} \\
\colorbox{blue}{\textbf{Qwen2.5 (3B $\rightarrow$ 0.5B)}}                & w/o  & w/ thinking & w/o & w/ thinking \\
\midrule
\textcolor{gray}{Student-0.5B} & \textcolor{gray}{0.386} & \textcolor{gray}{0.034} & \textcolor{gray}{0.166} & \textcolor{gray}{0.013} \\
\textcolor{gray}{Teacher-3B} & \textcolor{gray}{0.589} & \textcolor{gray}{0.560} & \textcolor{gray}{0.413} & \textcolor{gray}{0.357} \\
PPO & \colorbox{yellow}{0.547} & \colorbox{yellow}{0.581} & 0.258 & 0.242 \\
KD & 0.540 & 0.544 & \colorbox{green}{0.321} & \colorbox{yellow}{0.256} \\
\midrule
\textbf{\method} &  \colorbox{green}{0.565} & \colorbox{green}{0.593} & \colorbox{yellow}{0.312} &  \colorbox{green}{0.287} \\
\bottomrule
\end{tabular}
}
\caption{Source referencing and thinking performances on NQ and MuSiQue datasets.}
\label{tab:source_thinking}
\end{minipage}
\hfill
\begin{minipage}[t]{0.48\linewidth}
\centering
\resizebox{\linewidth}{!}{%
\begin{tabular}{lcccc}
\toprule
Models & \multicolumn{1}{c}{NQ (first hop)} & \multicolumn{2}{c}{MuSiQue (multi-hop)} \\
\colorbox{blue}{\textbf{Qwen2.5 (3B $\rightarrow$ 0.5B)}}                & Hit ratio  & Hit ratio & Search steps \\
\midrule
\textcolor{gray}{Student-0.5B}         &  \textcolor{gray}{0.004}      & \textcolor{gray}{0.052}        & \textcolor{gray}{3.86}         \\
\textcolor{gray}{Teacher-3B}           & \textcolor{gray}{0.682}       &  \textcolor{gray}{0.668}   &   \textcolor{gray}{1.60}      \\
PPO                  & \colorbox{green}{0.711}       &   0.568     &    1.68     \\
KD                   & 0.675       &  \colorbox{yellow}{0.570}      &   2.45      \\
\midrule
\textbf{\method}              & \colorbox{yellow}{0.682}       &     \colorbox{green}{0.583}   &     2.64    \\
\bottomrule
\end{tabular}
}
\caption{Query rewriting performance on NQ and thinking performance on MuSiQue datasets.}
\label{tab:hit}
\end{minipage}
\end{table*}

\subsection{Main Results ($\mathcal{Q}$1)}
\paragraph{Qwen 3B$\rightarrow$0.5B.}
Table~\ref{tab:qa_performance} shows the overall performance of different methods across seven QA benchmarks.
Our method consistently outperforms all baseline methods on most datasets and achieves the highest average score.
Remarkably, our method even surpasses the teacher model on three datasets,
\diffJ{as selective teacher guidance stabilizes training while RL enables policy improvements beyond teacher imitation.}
Among the on-policy methods that rely solely on SGOs, both PPO and GKD exhibit lower performance compared to off-policy distillation methods,
\diff{
due to the difficulty of the multi-turn agentic RAG task and the student’s near-zero initial performance, which makes SGOs highly noisy.
}
This result highlights the limitations of SGOs, which tend to be noisy and less informative than TGOs.
\final{Due to the severe capacity constraint of the compact model, SFT-based warm-start combined with RL does not lead to substantial additional improvements (0.275→0.289).
In contrast, our KD-based initialization distills the full soft output distribution of the teacher rather than only hard targets. The KD-based initialization alone already leads to stronger performance (0.298) than SFT-based initialization and RL.}
DistiLLM and TAID perform worse than standard KD.
In our setting, where the student model starts with extremely low performance, interpolating between the teacher and student distributions might have created noisy or misleading targets, resulting in weaker learning.

\diffJ{
\paragraph{Qwen 7B$\rightarrow$0.5B and Llama 8B$\rightarrow$1B.}
Table~\ref{tab:model_family} shows the average EM scores for models with a larger capacity gap (Qwen2.5 0.5B and 7B) and another model family (Llama3 1B and 8B). 
DGPO consistently outperforms both PPO and KD
across challenging capacity gaps (7--8B$\rightarrow$0.5--1B) and different model architectures (Qwen vs.\ Llama3).
While Qwen 3B$\rightarrow$0.5B slightly outperforms Qwen 7B$\rightarrow$0.5B due to a smaller capacity gap, DGPO effectively exploits compact model potential regardless of the teacher quality.
All results can
be found in Appendix~\ref{sec:detailed_results}.}

\begin{table*}[t]
\centering
\renewcommand{\arraystretch}{1.2} 
\resizebox{\textwidth}{!}{%
\begin{tabular}{l|ccc|cccccccc}
\toprule
Method  & \scalebox{1}[1]{init (KD)} & pipeline & \scalebox{1}[1]{KL penalty} & NQ & \scalebox{0.8}[1]{TriviaQA} & PopQA & \scalebox{0.8}[1]{HotpotQA} & 2wiki & \scalebox{0.8}[1]{MuSiQue} & \scalebox{0.8}[1]{Bamboogle} & Avg. \\
\midrule
\textbf{\method}   & \checkmark & \scalebox{0.9}[1]{KD $\rightarrow$ PPO} & selective  & 0.378 & 0.481 & 0.402  & 0.342  & 0.303 & 0.120  & 0.274  & 0.329  \\
\hdashline
(a) w/o \scalebox{0.9}[1]{cold-start initialization} & -- & \scalebox{0.9}[1]{KD $\rightarrow$ PPO} & selective & 0.370 & 0.465 & 0.394  & 0.330  & 0.299  & 0.117  & 0.266  & 0.320  \\
(b) w/o selective kl penalty &\checkmark & \scalebox{0.9}[1]{KD $\rightarrow$ PPO} & uniform & 0.362 & 0.464 & 0.394  & 0.323  & 0.306  & 0.114  & 0.234  & 0.314  \\
(c) w/o teacher guidance & \checkmark & \scalebox{0.9}[1]{KD $\rightarrow$ PPO} & -- & 0.353 & 0.455 & 0.384  & 0.316  & 0.287  & 0.098  & 0.250  & 0.306  \\
(d) \diff{invert pipeline order} & -- & \scalebox{0.9}[1]{PPO $\rightarrow$ KD} & -- & 0.320 & 0.426 & 0.371  & 0.287  & 0.282  & 0.084  & 0.234  & 0.286  \\

\bottomrule
\end{tabular}
}%
\caption{\diffJ{\textbf{Ablation study} on DGPO components. Results show the importance of cold-start initialization, selective KL penalty, teacher guidance during RL, and RL after KD initialization. See \Cref{tab:ablation_components} in \Cref{sec:ablation_baseline_setting} for more detailed configurations.}}
\label{tab:ablation}
\end{table*}

\subsection{ARCap -- Source Referencing ($\mathcal Q$2a)}
\paragraph{Setup.}
To isolate the capability of Source Referencing from other agentic behaviors, we evaluate the model's accuracy when provided only with the ground-truth supporting contexts (i.e., golden knowledge) as \colortag{brown}{information}, and forced to answer directly using the \colortag{teal}{answer} tag. 
For the MuSiQue dataset, which consists of multi-hop questions requiring multiple supporting documents, we concatenate all relevant ground-truth contexts and supply them as \colortag{brown}{information}. For the NQ dataset, we use the annotated long answer span as the input \colortag{brown}{information}. The final answer's correctness is measured using EM. 
\paragraph{Results.}
Table~\ref{tab:source_thinking} (w/o thinking column) shows the results for source referencing capability.
Our model achieves the highest score in extracting information from a single context on the NQ dataset. However, on the MuSiQue dataset, the KD model performs best. One possible explanation is that our RL phase may have over-optimized for simpler, single-step examples during RL, leading to suboptimal performance on complex multi-hop questions.

\subsection{ARCap -- Query Rewriting ($\mathcal{Q}$2b)}
\paragraph{Setup.}
To isolate the Query Rewriting capability from other agentic behaviors, we evaluate whether the initial search query formulated by the model can retrieve documents containing the correct answer, using the NQ dataset. 
As the evaluation metric, we adopt Hit ratio \cite{ma-etal-2023-query}, which measures whether at least one of the retrieved documents includes the correct answer.
\paragraph{Results.}
Table~\ref{tab:hit} (NQ column) shows the results for query rewriting.
Interestingly, the PPO model achieves the best performance, even surpassing the teacher model.
Our DGPO performs better than KD but reaches a similar hit ratio to the teacher.
This may be attributed to our training setup, which mixes both single-hop and multi-hop datasets.
Given the limited capacity of the student model, the PPO agent may have focused its exploration on simpler single-hop query writing tasks, rather than the more complex multi-hop reasoning required in other datasets.

\subsection{ARCap -- Thinking ($\mathcal {Q}$2c)}
\paragraph{Setup.}
To evaluate the Thinking capability, we assess \emph{how} and \emph{when} the model retrieves and integrates information during the reasoning process.  
\emph{(How:)} We provide the ground-truth contexts as \colortag{brown}{information} and allow the model to perform an additional \colortag{olive}{think} step immediately after \colortag{brown}{information} (i.e., the second \colortag{olive}{think} block in Table~\ref{tab:reasoning-example}). 
Note that such additional thinking was disallowed in the source referencing evaluation ($\mathcal Q$2a).
While further retrieval is technically unnecessary, the model is still allowed to perform additional search steps. 
\emph{(When:)} We allow multiple retrieval steps and examine whether the model can determine the necessity of additional searches based on intermediate results. In this case, we evaluate both the final Hit ratio and the average number of search steps taken as metrics of efficiency.

\begin{figure}[t]
  \centering
  \includegraphics[width=\linewidth]{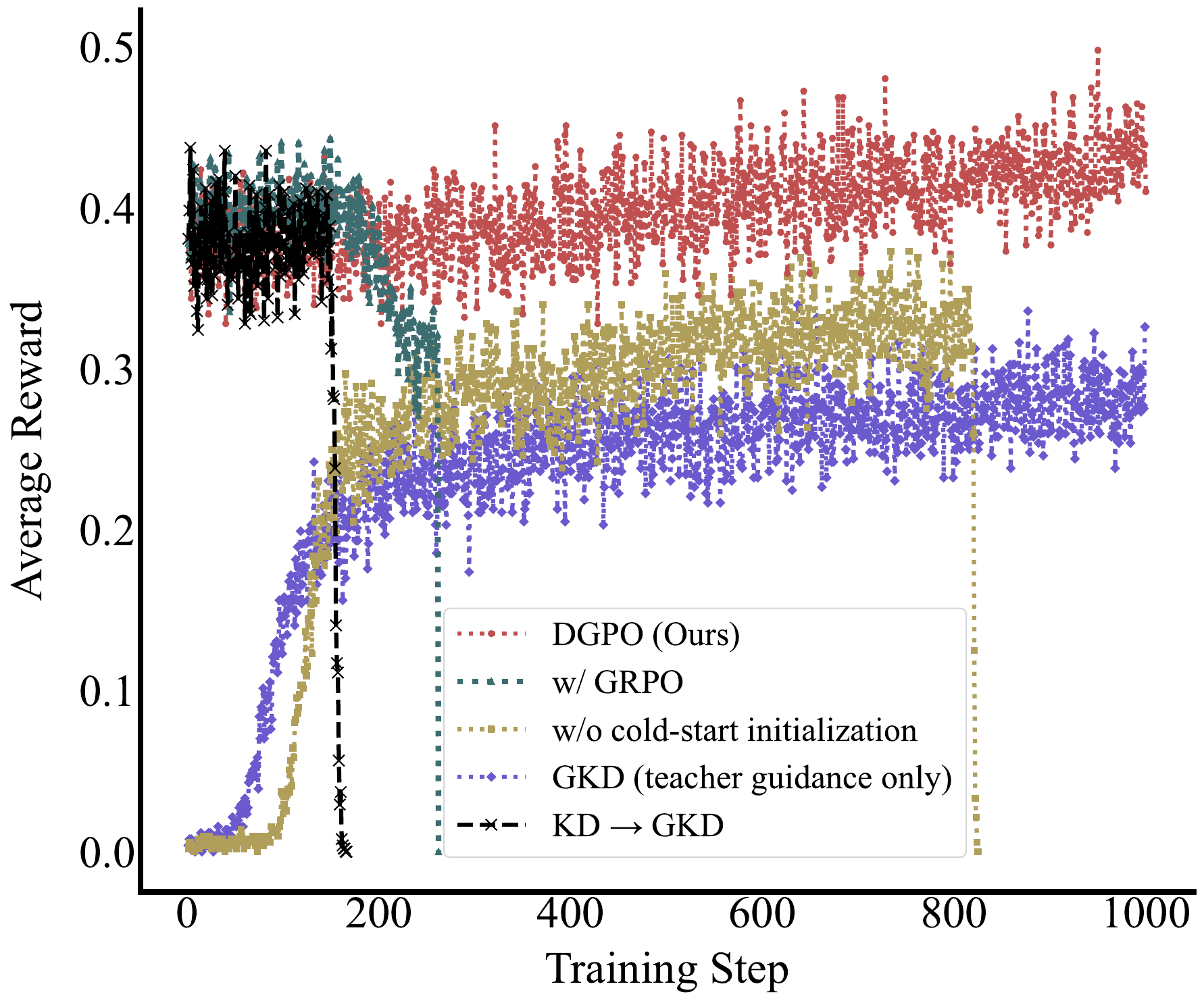}
  \caption{\diff{Training curves comparing \method\ and its ablations: (1) GRPO version; 
(2) without cold-start initialization; 
(3) GKD; and (4) KD→GKD. }
}
  \label{fig:curve}
\end{figure}

\paragraph{Results.}
As shown in Table~\ref{tab:source_thinking} (w/ thinking column), many models, including the teacher, exhibit performance degradation when additional \colortag{olive}{think} steps are introduced. This suggests that under our smaller model setting, deliberate reasoning through thinking is not crucial for information extraction. Only the RL models improve on the NQ dataset. They may have learned to use thinking to double-check their answers for simpler setting.

As shown in Table~\ref{tab:hit} (MuSiQue column), while the PPO model performs well in the first retrieval step, our method achieves the highest score for more complex multi-hop reasoning. 
To achieve higher hit ratios, the distilled model tends to take more search steps.
Compared to the teacher, which achieves strong performance with fewer steps due to its larger capacity, our method enables the student to compensate by exploring more extensively.

\subsection{Ablation Study ($\mathcal Q$3)}
Table~\ref{tab:ablation} presents the results of our ablation study.
\diff {(a)} \diff{w/o cold-start} initialization by KD, the performance drop is relatively small; however, training becomes unstable and collapses around step 800, so we report the score just before the collapse. 
\diff{(b) w/o selective KL penalty applies KL regularization uniformly across all trajectories, regardless of whether the student's attempt is correct or incorrect.
(c) w/o teacher guidance denotes KD initialization followed by standard PPO without KL regularization during RL.
Both variants (b) and (c) result in performance degradation for our method.
(d) Reversing the order \diff{(PPO before KD) causes} substantial performance loss.
These results confirm that all proposed components are essential: KD initialization prevents collapse, pipeline KD$\rightarrow$PPO with selective KL penalty is crucial.
}

\subsection{Training Dynamics ($\mathcal{Q}$4)}
Figure~\ref{fig:curve} illustrates
the training stability of \method and its variants across different RL algorithms and initialization strategies.
\method maintained a stable training curve
beyond 1000 steps, achieving the best overall performance. 
However, (1) replacing PPO with GRPO leads to an early collapse during RL. 
Even with KD initialization and teacher guidance, GRPO remains unstable for compact models. 
(2) When removing KD initialization from our model, 
training remains more stable until 800 steps compared to the standard PPO but collapses at around 800 steps.
(3) Using GKD, i.e., teacher guidance only, results in stable learning; however, the absence of self-exploration in RL leads to worse performance.
(4) When KD-based initialization is further combined with GKD, training collapses  due to the excessive constraints imposed by the teacher.

\section{Conclusion}
\label{sec:conclusion}

We propose Distillation-Guided Policy Optimization (DGPO), a novel RL framework that overcomes the core challenge of poor SGOs in compact models via cold-start initialization and selective teacher guidance. 
DGPO transforms the reference model from a passive regularizer to an active guidance mechanism, enabling performance improvements rather than merely preventing degradation. 
Our two-phase approach achieves consistent improvements without complex scheduling.
Beyond end-to-end gains, our ARCap-based analysis provides a fine-grained breakdown of how \method improves agentic behavior, highlighting its strengths across dimensions such as source referencing, query rewriting, and multi-hop reasoning.
\paragraph{Can compact language models search like agents?}
Our findings suggest \textbf{yes}. Starting from a 0.5B model with minimal performance (0.006), DGPO achieves a 55× improvement (0.329), approaching the 3B teacher's performance (0.353). Remarkably, our student model even surpasses the teacher on several datasets.
Given that 0.5B models can run efficiently on CPUs, our method democratizes access to search agents across computing resource-constrained devices like laptops and smartphones, opening possibilities for more practical agentic RAG deployment.
\diff{As a foundational study on enabling agentic RAG in compact models, we focus on QA tasks for comprehensive evaluation. Future work will extend this approach to diverse tasks requiring agentic reasoning.
}

\paragraph{Acknowledgements}
This work was supported by the “Development Acceleration Use” program of ABCI 3.0, which is provided by AIST and AIST Solutions, and JST AIP Acceleration Research, Japan, Grant Number JPMJCR23U2 and JST PRESTO, Japan, Grant Number JPMJPR2518.

\clearpage 
\newpage
\section*{Limitations}
\label{sec:limitation}
\diff{Our experiments are restricted to Qwen2.5 (3B$\rightarrow$0.5B, 7B$\rightarrow$0.5B) and Llama3 (8B$\rightarrow$1B) model families.
Given the rapid advancement of LLMs, comprehensive evaluation across all available models is impractical within current research timelines.
Due to computational limitations, we restrict our investigation to student models of 0.5--1B parameters and teacher models up to 8B parameters.
While larger teacher models are available,
this work specifically targets compact models for computing resource-constrained environments, making exploration of massive teacher models beyond both our computational capacity and research scope.}

As stated in Section~\ref{sec:exp}, while our model achieves strong overall performance, 
\diff{optimization across all capacity dimensions remains an open challenge.}
\diff{We believe that our ARCap analysis framework and proposed DGPO approach provide essential foundations for enabling compact models to acquire sophisticated agentic behaviors.}

\final{
Although the use of a teacher model introduces additional overhead, under our setup where a 3B teacher model is used only for inference during RL while a 0.5B student model is fully trained, it increases the overall training time by 9.5\%.
This overhead is relatively small given the stability and performance improvements achieved. }

\final{
Our primary motivation is to study distillation for compact models, specifically in the context of agentic RAG. We agree that transferring to other tasks is important. Extending beyond agentic RAG remains valuable future work.}

\bibliography{custom}
\appendix
\section*{Appendix}
\label{sec:appendix}
\begin{table}[h]
\small
    \centering
    \begin{tabular}{p{7cm}}
        \toprule
        \textbf{System Template for qwen2.5 series} \\
        \midrule
        You are Qwen, created by Alibaba Cloud. You are a helpful assistant. \\
        \bottomrule
        \noalign{\vskip 2mm}
        \toprule
        \textbf{Instruction Template} \\
        \midrule
        Answer the given question. You must conduct reasoning inside \colortag{olive}{think} and \colortag{olive}{/think}
first every time you get new information. After reasoning, if you find you lack some
knowledge, you can call a search engine by \colortag{sblue}{search} query \colortag{sblue}{/search}, and it will
return the top searched results between \colortag{brown}{information} and \colortag{brown}{/information}. You
can search as many times as you want. If you find no further external knowledge
needed, you can directly provide the answer inside \colortag{teal}{answer} and \colortag{teal}{/answer} without
detailed illustrations. For example, \colortag{teal}{answer} xxx \colortag{teal}{/answer}. Question: \textcolor{orange}{\texttt{question}}.\\
        \bottomrule
    \end{tabular}
    \caption{System and instruvtion template for agentic RAG. \textcolor{orange}{\texttt{question}} is replaced with the specific question during training and inference.}\label{tab:template}
\end{table}

\begin{table}[h]
\centering
\resizebox{\linewidth}{!}{  
\begin{tabular}{|l|l|c}  
\toprule  
Config & Parameter & Value \\  
\midrule   
RL parameters & Total training steps & 1000 \\ 
&Batch size & 512 \\   
&KL divergence coefficient $\beta$ & 0.001 \\ 
&Maximum prompt length & 4096 \\  
&Maximum response length & 500 \\  
&Maximum conversation turns & 4 \\  
&Top-k retrieved documents & 3 \\  
&Actor learning rate & 1e-6 \\  
&Critic learning rate & 1e-5 \\
\midrule
KD parameters & Tortal epochs & 5 \\
(initialization) & Batch size  & 64 \\
&Learning rate & 1e-4 \\
&KL divergence ratio $\lambda$ & 1.0 \\
\midrule   
DistiLLM~\cite{distillm} & Skew KLD target weight & 0.1 \\ 
\midrule   
TAID~\cite{shing2025taid} & $t_{start}$ & 0.4 \\   
&$t_{end}$ & 1.0 \\  
&Updating interpolation ($\alpha$) & 5e-4 \\ 
&Momentum coefficient ($\beta$) & 0.99 \\   
\bottomrule  
\end{tabular}  
}
  \caption{\diff{Parameters for \method and baselines.}}
  \label{tab:rl_parameter}
\end{table}

\begin{table*}[h]
\centering
\small
\begin{tabular}{l r r l}
\toprule
Dataset & Training samples & Test samples & License \\
\midrule
Natural Questions (NQ) \citep{nq}              & 79{,}168 & 3{,}610  & CC BY-SA 3.0 \\
TriviaQA \citep{triviaqa}                      & --       & 11{,}313 & Apache-2.0 \\
PopQA \citep{popqa}                            & --       & 14{,}267 & MIT \\
HotpotQA \citep{yang2018hotpotqa}              & 90{,}447 & 7{,}405  & CC BY-SA 4.0 \\
2WikiMultiHopQA \citep{2wiki}                  & --       & 12{,}576 & Apache-2.0 \\
MuSiQue \citep{musique}                        & --       & 2{,}417  & CC BY 4.0 \\
Bamboogle \citep{bamboogle}                    & --       & 125      & MIT \\
\bottomrule
\end{tabular}
\caption{Statistics of training and test datasets.}
\label{tab:data}
\end{table*}

\begin{table*}[h]
    \centering
    \footnotesize
    \begin{tabular}{l|c|c|c|c|c|c}
        \toprule
        \textbf{Setting} & Results & \rotatebox{90}{KD}  \rotatebox{90}{(initialization)} & \rotatebox{90}{PPO Loss} & \rotatebox{90}{GRPO Loss} & \rotatebox{90}{Selective}  \rotatebox{90}{KL penalty} & \rotatebox{90}{Uniform}  \rotatebox{90}{KL penalty} \\
        \midrule
        DGPO & Tab. \ref{tab:qa_performance} & \checkmark & \checkmark & & \checkmark & \\
        \midrule
        w/ GRPO & Fig. \ref{fig:curve} & \checkmark & & \checkmark & \checkmark & \\
        w/o cold-start initialization & Tab. \ref{tab:ablation} & & \checkmark & & \checkmark & \\
        w/o selective KL penalty (uniform KL penalty) & Tab. \ref{tab:ablation} & \checkmark & \checkmark & & & \checkmark \\
        w/o teacher guidance (KD→PPO) & Tab. \ref{tab:ablation} & \checkmark & \checkmark & & & \\
        invert pipeline order (PPO→KD) & Tab. \ref{tab:ablation} & \checkmark & \checkmark & & & \\
        KD→GKD & Fig. \ref{fig:curve} & \checkmark & & & & \checkmark \\
        PPO \citep{searchr1} & Tab. \ref{tab:qa_performance} & & \checkmark & & & \\
        KD \citep{kd}& Tab. \ref{tab:qa_performance} & \checkmark & & & & \\
        GKD \citep{gkd}& Tab. \ref{tab:qa_performance} & & & & & \checkmark \\
        \bottomrule
    \end{tabular}
    \caption{\diff{Ablation and baseline settings and their components.}}
    \label{tab:ablation_components}
\end{table*}

\section{\move{RL for Agentic RAG}}
\label{app:RL-RAG}

\move{
We ground the reinforcement learning framework on the skeletal formalization of Search-R1~\cite{searchr1}, which is one of the state-of-the-art agentic RAG frameworks.
We model the agentic search process as a sequential decision-making problem where the LLM agent must learn to coordinate reasoning and retrieval operations.
At each step, the agent can either generate text to advance its reasoning or issue queries to the external search engine $\mathcal R$ to gather additional information.
\paragraph{Learning Objective.}
The Reinforcement Learning for agentic RAG framework is formulated as:
\begin{align}
    \max_{\pi_\theta} &\mathbb{E}_{x \sim \mathcal{D}, y \sim \pi_{\theta}(\cdot \mid x; \se)} 
\left[ r_{\phi}(x, y) \right] \nonumber
\\
&- \beta \mathbb{D}_{\text{KL}} \left[ \pi_{\theta}(y \mid x; \se) \,||\, \pi_{\text{ref}}(y \mid x; \se) \right],
\end{align}
where
$\pi_\theta$ denotes the trainable agent policy that generates action trajectories $y$ conditioned on the input user question $\bm x$ and an external retrieval system $\mathcal R$.
The reward function $r(\bm x,\bm y)$ evaluates accuracies of generated answers.
The KL-divergence term with coefficient $\beta$ provides regularization against the frozen reference policy $\pi_{\text{ref}}$.
}

\section{Implementation Details}
\label{app:implementation}
\subsection{Token Masking} 
\label{sec:token-masking}
Following prior work~\cite{searchr1}, we employ token masking during training. 
\cref{eq:ppo}, $\mathbbm1 (y_t)$ is the loss-masking operator defined as,
\begin{equation}
\mathbbm{1}(y_t) = \begin{cases}
1 & \text{if } y_t \in \{ \texttt{LLM-generated tokens} \} \\
0 & \text{if } y_t \in \{\texttt{external tokens}\}\,.
\end{cases}
\end{equation}
In agentic RAG, the token sequence contains both LLM agent-generated tokens (\colortag{sblue}{search}, \colortag{olive}{think}, and \colortag{teal}{answer}) and externally retrieved content from the search system $\mathcal R$ (\colortag{brown}{information}).
Computing gradients over retrieved tokens is counterproductive, as it encourages the model to learn how to generate external content rather than focusing on the core agentic capabilities of when and how to search. 
To prevent this misallocation of model capacity and stabilize training, we apply loss masking to retrieved tokens and documents, ensuring optimization focuses solely on agent-generated content.

\subsection{Prompt Template}
We used the system template for Qwen2.5 series and the instruction template following \citet{searchr1}. Table~\ref{tab:template} shows these templates.

\begin{table*}[h]
\centering
\small
\resizebox{\textwidth}{!}{%
\begin{tabular}{l|ccc|cccc|c}
\toprule
\colorbox{blue}{\textbf{Qwen 2.5 (7B $\rightarrow$ 0.5B)}} & NQ & TriviaQA & PopQA & HotpotQA & 2wiki & MuSiQue & Bamboogle & Avg. \\
\midrule
\textcolor{gray}{Student-0.5B}        & \textcolor{gray}{0.004} & \textcolor{gray}{0.006} & \textcolor{gray}{0.007} & \textcolor{gray}{0.007} & \textcolor{gray}{0.015} & \textcolor{gray}{0.000} & \textcolor{gray}{0.000} & \textcolor{gray}{0.006} \\
\textcolor{gray}{Teacher-7B}          & \textcolor{gray}{0.393} & \textcolor{gray}{0.610} & \textcolor{gray}{0.397} & \textcolor{gray}{0.370} & \textcolor{gray}{0.414} & \textcolor{gray}{0.146} & \textcolor{gray}{0.368} & \textcolor{gray}{0.385} \\
\hdashline
PPO \cite{searchr1} & 0.306 & \colorbox{yellow}{0.444} & \colorbox{yellow}{0.379} & 0.205 & 0.218 & 0.041 & 0.073 & 0.238 \\
KD \cite{kd}               &  \colorbox{yellow}{0.338}& 0.428&  0.371&  \colorbox{yellow}{0.288}& \colorbox{yellow}{0.223} & \colorbox{yellow}{0.100} & \colorbox{yellow}{0.210} & \colorbox{yellow}{0.280} \\
\midrule
\textbf{\method} (ours)               & \colorbox{green!80}{0.371} &\colorbox{green!80}{0.474} &\colorbox{green!80}{0.396}  &\colorbox{green!80}{0.334}  &\colorbox{green!80}{0.257}  &\colorbox{green!80}{0.113}  &\colorbox{green!80}{0.315}  &\colorbox{green!80}{0.323}  \\
\bottomrule
\end{tabular}
}  
\caption{\diffJ{\textbf{Qwen 2.5 (7B $\rightarrow$ 0.5B) results across different methods and QA benchmarks}. The best and second-best results are highlighted in \colorbox{green!80}{green} and \colorbox{yellow}{yellow}, respectively.}}
\label{tab:qwen7b_performance}
\end{table*}

\begin{table*}[h]
\centering
\small
\resizebox{\textwidth}{!}{%
\begin{tabular}{l|ccc|cccc|c}
\toprule
\colorbox{red}{\textbf{Llama 3 (8B $\rightarrow$ 1B)}} & NQ & TriviaQA & PopQA & HotpotQA & 2wiki & MuSiQue & Bamboogle & Avg. \\
\midrule
\textcolor{gray}{Student-1B}        & \textcolor{gray}{0.052} & \textcolor{gray}{0.080} & \textcolor{gray}{0.044} & \textcolor{gray}{0.027} & \textcolor{gray}{0.042} & \textcolor{gray}{0.001} & \textcolor{gray}{0.024} & \textcolor{gray}{0.039} \\
\textcolor{gray}{Teacher-8B}          & \textcolor{gray}{0.475} & \textcolor{gray}{0.647} & \textcolor{gray}{0.448} & \textcolor{gray}{0.427} & \textcolor{gray}{0.443} & \textcolor{gray}{0.179} & \textcolor{gray}{0.444} & \textcolor{gray}{0.438} \\
\hdashline
PPO \cite{searchr1} & 0.354 & 0.499 & 0.394 & 0.222 & 0.181 & 0.037 & 0.065 & 0.250 \\
KD \cite{kd}               &  \colorbox{yellow}{0.406}& \colorbox{yellow}{0.508}&  \colorbox{yellow}{0.405}&  \colorbox{yellow}{0.369}& \colorbox{yellow}{0.355} & \colorbox{yellow}{0.119} & \colorbox{yellow}{0.266} & \colorbox{yellow}{0.347} \\
\midrule
\textbf{\method} (ours)               & \colorbox{green!80}{0.448} &\colorbox{green!80}{0.553} &\colorbox{green!80}{0.437}  &\colorbox{green!80}{0.412}  &\colorbox{green!80}{0.379}  &\colorbox{green!80}{0.155}  &\colorbox{green!80}{0.339}  &\colorbox{green!80}{0.389}  \\
\bottomrule
\end{tabular}
}  
\caption{\diffJ{\textbf{Llama 3 (8B $\rightarrow$ 1B) results across different methods and QA benchmarks}. The best and second-best results are highlighted in \colorbox{green!80}{green} and \colorbox{yellow}{yellow}, respectively.}}
\label{tab:llama8b_performance}
\end{table*}

\subsection{Training Details}
\label{sec:training-settings}
On-policy distillation or RL methods were trained for up to 1000 steps.
However, PPO training with a small model is inherently unstable; thus, we report the results at step 200, before training collapse.
All models were initialized from the same pretrained checkpoints and trained once.
Training took approximately one day on 8×H200 GPUs.
The hyperparameters and libraries used for implementation followed those of prior work \cite{searchr1,shing2025taid}.
Table \ref{tab:rl_parameter} shows training parameters.

\subsection{Dataset Details}
\label{app:dataset-details}
We used preprocessed seven QA datasets following \citet{searchr1}.
Table~\ref{tab:data} shows dataset statistics.
These datasets are originally designed for QA tasks, and our use aligns with their intended purpose.

\section{\diff{Ablation and Baseline Settings}}
\label{sec:ablation_baseline_setting}
\diff{Table~\ref{tab:ablation_components} summarizes the ablation and baseline settings used in our study, indicating which components (e.g., KD, PPO loss, GRPO loss, selective or uniform KL penalties) are included in each variant, along with references to the corresponding figures or tables where results are reported.}

\section{\diffJ{Evaluation across Different Model Families and Larger Capacity Gaps}}
\label{sec:detailed_results}
\diffJ{
We evaluate our method across different model families and larger capacity gaps using state-of-the-art compact models (0.5--1B students). Table~\ref{tab:qwen7b_performance} shows Qwen2.5 (7B $\rightarrow$ 0.5B) results and Table~\ref{tab:llama8b_performance} shows Llama 3 (8B $\rightarrow$ 1B) results. Both configurations demonstrate DGPO's consistent superiority over baseline methods, confirming broad applicability across diverse architectures and capacity gaps.}

\end{document}